\documentclass[runningheads]{llncs}

\usepackage{eccv}

\usepackage{eccvabbrv}

\usepackage{graphicx}
\usepackage{booktabs}

\usepackage[accsupp]{axessibility}

\usepackage{graphicx}
\usepackage{amsmath}
\usepackage{amssymb}
\usepackage{booktabs}
\usepackage{enumitem}
\usepackage{multirow}
\usepackage{makecell}
\usepackage[ruled,linesnumbered]{algorithm2e}
\usepackage{algorithmicx}
\usepackage{pifont}
\newcommand{\cmark}{\ding{51}}
\newcommand{\xmark}{\ding{55}}
\newsavebox\CBox
\def\textBF#1{\sbox\CBox{#1}\resizebox{\wd\CBox}{\ht\CBox}{\textbf{#1}}}

\usepackage{hyperref}

\usepackage{orcidlink}

\begin{document}

\title{Canny2Palm: Realistic and Controllable Palmprint Generation for Large-scale Pre-training} 

\titlerunning{Canny2Palm: Realistic and Controllable Palmprint Generation}

\author{Xingzeng Lan\orcidlink{0009-0003-4163-0084}, Xing Duan\orcidlink{0000-0002-9423-8268}, Chen Chen\orcidlink{0009-0006-2823-7108}, Weiyu Lin\orcidlink{0009-0009-9846-9125}, Bo Wang\orcidlink{0009-0001-6058-2159}}

\authorrunning{X. Lan et al.}

\institute{Deptrum\\
\email{\{xingzeng.lan, xing.duan, chen.chen, weiyu.lin, bo.wang\}@deptrum.com}\\}

\maketitle

\begin{abstract}
  Palmprint recognition is a secure and privacy-friendly method of biometric identification.
  One of the major challenges to improve palmprint recognition accuracy is the scarcity of palmprint data. 
  Recently, a popular line of research revolves around the synthesis of virtual palmprints for large-scale pre-training purposes.
  In this paper, we propose a novel synthesis method named Canny2Palm that extracts palm textures with Canny edge detector and uses them to condition a Pix2Pix network for realistic palmprint generation.
  By re-assembling palmprint textures from different identities, we are able to create new identities by seeding the generator with new assemblies.
  Canny2Palm not only synthesizes realistic data following the distribution of real palmprints but also enables controllable diversity to generate large-scale new identities.
  On open-set palmprint recognition benchmarks, models pre-trained with Canny2Palm synthetic data outperform the state-of-the-art with up to 7.2\% higher identification accuracy.
  Moreover, the performance of models pre-trained with Canny2Palm continues to improve given 10,000 synthetic IDs while those with existing methods already saturate, demonstrating the potential of our method for large-scale pre-training.
  
\end{abstract}

\section{Introduction}
\label{sec:intro}

The deployment of face recognition in the public has raised privacy concerns over the years due to the traceability of facial images acquired without consent.
Additionally, face recognition accuracy degrades when the face is partially covered by masks, which is common amidst the COVID era.
Palmprint recognition emerges as an alternative and avoids these two challenges.
Only when the subjects cooperatively present their hands, a palmprint recognition device can acquire the palmprint data.  
Owing to its advantages in privacy, accuracy and convenience, palmprint recognition has received significant attention in both academia \cite{zhong2019decade} and industry \cite{zhang2019pay}.

In contrast to the extensive and publicly available face datasets, palmprint datasets are limited and small.
The largest public palmprint recognition dataset known to us contains only thousands of identities \cite{kumar2018toward}. 
In comparison, face recognition datasets have expanded to millions of identities \cite{zhu2021webface260m}. 
The insufficient amount of palmprint data limits the performance of deep learning models.

Recently, the emergence of synthetic data shows promises to supplement palmprint data scarcity \cite{salazar2022towards}.
B$\rm\acute{e}$zierPalm \cite{zhao2022bezierpalm} represents palmprints with B$\rm\acute{e}$zier curves and synthesizes new identities by randomly sampling B$\rm\acute{e}$zier parameters.
Thanks to the explicit parameters, B$\rm\acute{e}$zierPalm can control the ID uniqueness and diversity.
However, B$\rm\acute{e}$zier curves are only a handcrafted first-order approximation to palmprints which leads to a domain gap between the synthetic and real palmprints.
Synthetic Style-based Palm Vein Database (Synthetic-sPVDB) \cite{salazar2021generating} adopts a StyleGAN model to learn the distribution of real palmprints and generate new identities with random noise.
This approach mitigates the domain gap issue.
Because of the uncontrollable seeds fed to the generator, a uniqueness classifier is required to classify generated palmprints into different IDs.
The classification accuracy limits the upper-bound of the synthetic data.
The trade-off between controllability and realisticity is a challenge to existing methods.

In this paper, we propose Canny2Palm, a method to synthesize realistic palmprints for new identities based on real palmprint distribution with full control of uniqueness and diversity.
Canny2Palm extracts palm texture with the Canny edge detector \cite{canny1986computational} and use it to condition the generator to reproduce the original palmprints.
This strategy ensures the model learns directly from the real data distribution while being controlled by features extracted from real data.
To generate new identities, Canny2Palm reassembles palm texture patches as the seed to generator.
We devise a reassembly mechanism to maximize inter-class distinction and intra-class diversity.
The performance of palmprint recognition models pre-trained with Canny2Palm synthetic data exhibit significant improvement.

\begin{figure}[t]
  \centering  
  \begin{subfigure}{0.18\linewidth}
    \centering
    \includegraphics[width=0.8\linewidth]{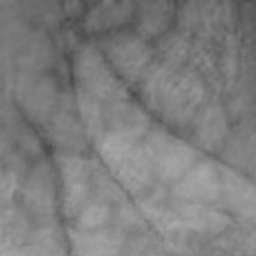}
    \caption{NSPVDB}
    \label{fig:synthetic_example-sub1}
  \end{subfigure}
  \begin{subfigure}{0.18\linewidth}
    \centering
    \includegraphics[width=0.8\linewidth]{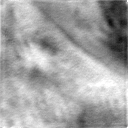}
    \caption{Syn-sPVDB}
    \label{fig:synthetic_example-sub2}
  \end{subfigure}
  \begin{subfigure}{0.18\linewidth}
    \centering
    {\includegraphics[width=0.8\linewidth]{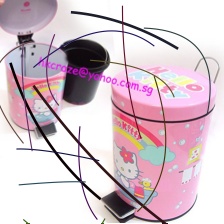}}
    \caption{B$\rm\acute{e}$zierPalm}
    \label{fig:synthetic_example-sub3}
  \end{subfigure}
  \begin{subfigure}{0.18\linewidth}
    \centering
    {\includegraphics[width=0.8\linewidth]{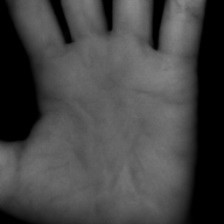}}
    \caption{Ours}
    \label{fig:synthetic_example-sub4}
  \end{subfigure}
  \begin{subfigure}{0.18\linewidth}
    \centering
    {\includegraphics[width=0.8\linewidth]{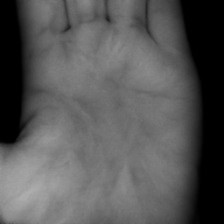}}
    \caption{Ours}
    \label{fig:synthetic_example-sub5}
  \end{subfigure}
  \caption{Example images of different synthetic datasets. (a) is the reproduced venous plexus of the palm based on the patterns of the Physarum; (b) is the vein texture generated by StyleGAN; (c) is the simulation of palmprint using B$\rm\acute{e}$zier curves, with a random background from ImageNet dataset; (d) and (e) are palmprint images with different gestures from the same identity based on our approach.}
  \label{fig:synthetic_example}
\end{figure}

The main advantages of the proposed method are summarized as follows:
\noindent \textbf{Controllable inter-class distinction}. Each combination of Canny blocks from different identities constructs a unique seed to the generator. We introduced two guidelines to preserve the differences between identities, which guarantees inter-class distinction without an extra uniqueness classifier as in \cite{salazar2021generating,salazar2021automatic,shen2023rpg}.

\noindent \textbf{Controllable intra-class diversity}. Canny2Palm is able to generate multiple samples for a synthetic identity by combining Canny textures outside the ROI region. The generated samples exhibit diverse hand features, surpassing the limitation of one sample per ID as in \cite{salazar2021generating,salazar2021automatic}.

\noindent\textbf{Realisticity}.
Canny2Palm generates realistic palmprints by training a Pix2Pix network on the real palmprint data.
This reduces the domain gap between synthetic and real data and enables pre-training to achieve higher accuracy.

\noindent \textbf{Accuracy and scalability}. Palmprint recognition models pre-trained on samples synthesized by Canny2Palm outperform the state-of-the-art on all benchmarks. When the number of synthesized IDs increases, the performance of the pre-trained model saturates significantly slower with Canny2Palm samples.

\section{Related Works}
\label{sec:related_work}

In palmprint recognition, models typically learn a mapping from the palm images to discriminative features, characterized by small intra-class and large inter-class distances \cite{deng2019arcface}.
Training a large model with insufficient data in high-dimensional space leads to overfitting due to the excessive model capacity.
In particular, the number of identities is crucial for metric learning.
To address the issue of data scarcity, several studies \cite{salazar2021generating, salazar2021automatic, zhao2022bezierpalm, shen2023rpg} investigated the potential of data synthesis.

\cite{salazar2021generating} proposed a Generative Adversarial Network \cite{goodfellow2014generative} (GAN)-based method to generate the textures of real palm with random noise vector $z$. 
\cite{salazar2021automatic} presents a nature-based approach to synthesize the venous plexus of the palm based on the growth pattern of the Physarum polycephalum. 
\cite{shen2023rpg} proposed a realistic pseudo-palmprint generation model by feeding random noise vector and B$\rm\acute{e}$zier curves into a conditional generator.
All of these studies employ random input or random conditions to generate images, and a uniqueness classifier is required to classify generated palmprints into different IDs.
\cite{zhao2022bezierpalm} proposed a controllable synthesis method using B$\rm\acute{e}$zier curves to simulate palmprints. This approach successfully guarantees intra-class diversity, but as the number of identities increases, the inter-class distinction tends to diminish. Additionally, the synthesized image primarily consists of colorful lines with random backgrounds, which deviates significantly from the real palm textures.

In this paper, we propose a novel GAN-based synthesis method using Pix2Pix \cite{isola2017image}, aiming to generate realistic samples with controllable intra-class diversity and clear inter-class distinction for palmprint recognition.

\section{Methodology}
\label{sec:methodology}

\subsection{Generate Palmprints from Canny Textures}
\label{sec:reproduce}

\begin{figure}
  \centering  
  \includegraphics[width=1.0\linewidth]{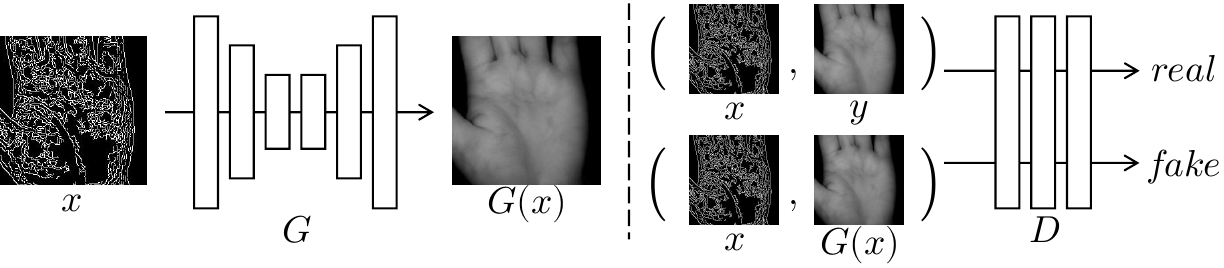}
  \caption{The training procedure of the palmprint generator. The generator produces realistic palmprints based on the palm texture, while the discriminator tries to classify between the real and fake pairs.}
  \label{fig:Pix2Pix_procedure}
\end{figure}

To train a generative model with control of the identity, we use the Pix2Pix network, a special architecture of the conditional GAN (cGAN) \cite{mirza2014conditional}.
\cref{fig:Pix2Pix_procedure} illustrates the training procedure of the generative model.
The cGAN network learns the mapping with a condition vector $x$ and a noise $z$, \ie, $G:\{x,z\}\rightarrow y$.
Pix2Pix discards the noise $z$ in order to make the generation process more controllable, since the condition vector $x$ is sufficiently complex.
The generation process $G:x\rightarrow y$ can be treated as an image-to-image translation.

\begin{figure}[b]
  \centering 
  \captionsetup[subfigure]{justification=centering}
  \begin{subfigure}{0.25\linewidth}
    \centering
    \includegraphics[width=0.8\linewidth]{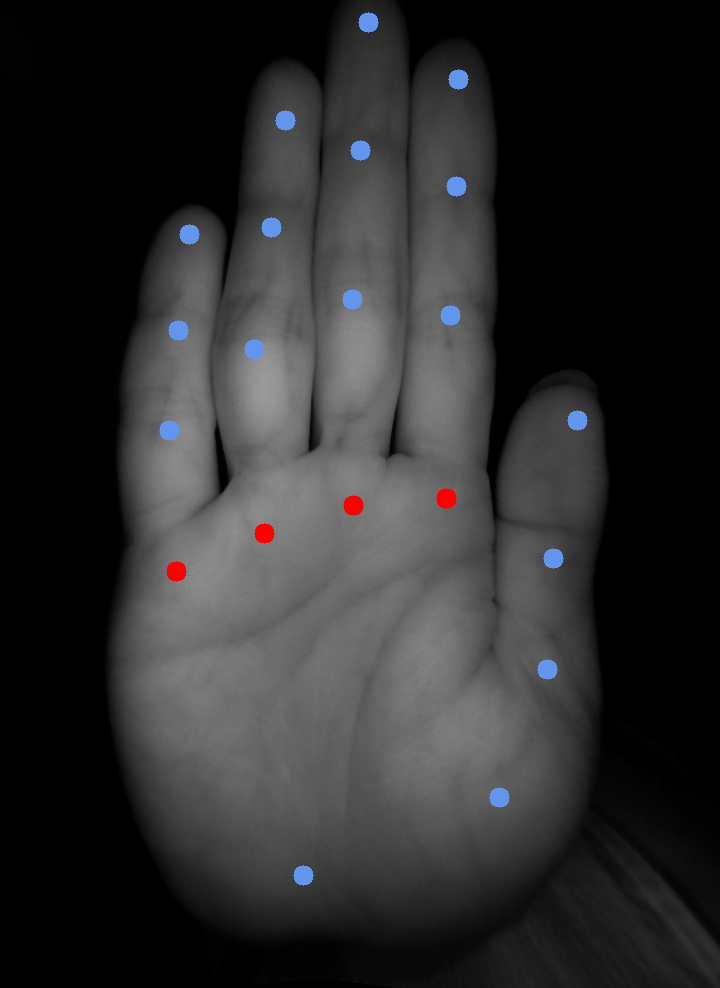}
    \caption{Original image \\ and hand key-points}
    \label{fig:alignment_canny-sub1}
  \end{subfigure}
  \begin{subfigure}{0.25\linewidth}
    \centering
    \raisebox{0.14\linewidth}
    {\includegraphics[width=0.8\linewidth]{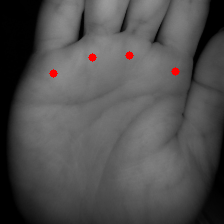}}
    \caption{Normalized palmprint}
    \label{fig:alignment_canny-sub2}
  \end{subfigure}
  \begin{subfigure}{0.25\linewidth}
    \centering
    \raisebox{0.14\linewidth}
    {\includegraphics[width=0.8\linewidth]{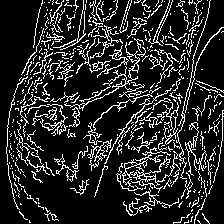}}
    \caption{Extracted palmprint texture}
    \label{fig:alignment_canny-sub3}
  \end{subfigure}
  \caption{An illustration of the palm normalization and Canny edge extraction. (a) demonstrates 21 standard hand key-points; (b) is the normalized image by affine transformation with four reference joint points in red; (c) is the texture extracted by Canny.}
  \label{fig:alignment_canny}
\end{figure}

The most significant features of palmprints for recognition are the palm texture, \ie, wrinkles and veins, especially those deep and principal lines.
Hence, we choose palm texture as the condition $x$ for the Pix2Pix generator.
In particular, we employ the Canny edge detector with a tiny low-threshold to extract redundant textures.
Considering the capacity of neural networks, we believe that redundant textures are preferable to fewer textures, especially when some palmprint images are unclear.
To guarantee the spatial consistency of features, palmprint images are normalized according to key-points before feeding into the model as illustrated in \cref{fig:alignment_canny-sub1} and \cref{fig:alignment_canny-sub2}.

As a straightforward embedding method, texture extraction acquires richer complexity thanks to various data sources.
Trainable methods such as VAE \cite{kingma2013auto} may be limited by the distribution of limited training data.
Moreover, extracting textures directly from real data avoids artificial bias as in \cite{zhao2022bezierpalm, salazar2021automatic} and ensures both realisticity and complexity.

The loss function of the network is defined as 
\begin{equation}
  \begin{aligned}
  \mathcal{L}_{Pix2Pix}(G, D) &= \mathbb{E}_{x,y}[\log{D(x,y)}] + \mathbb{E}_{x}[\log{(1-D(x, G(x)))}]
  \end{aligned}
  \label{eq:loss_GAN}
\end{equation}

In addition, a $L_1$ loss is added to the objective to measure the similarity between the generated palmprint $G(x)$ and the original image $y$.
We choose $L_1$ distance rather than $L_2$ because $L_1$ encourages less blurring \cite{isola2017image}:

\begin{equation}
    \mathcal{L}_{1}(G)=\Vert y-G(x)\Vert_{1}
\label{eq:loss_l1}
\end{equation}

The final objective is defined as
\begin{equation}
    \min_{G}\max_{D}\mathcal{L}(G, D)= \mathcal{L}_{Pix2Pix}(G, D)+\lambda\mathcal{L}_{1}(G)
\label{eq:loss_final}
\end{equation}
where $\lambda$ is a parameter balancing Pix2Pix loss and $L_1$ loss. 

\subsection{Synthesize New Identities with Controllable Intra-class Diversity}
\label{sec:synthesize-new}

With a well-trained generator, we are able to generate palmprints from any Canny edge image.
A unique input corresponds to a unique identity.
Palmprint recognition models pay most of the attention to an ROI of the palm.
Based on this observation, we can synthesize new identities by assembling new ROIs.

\begin{figure}[t]
  \centering  
  \captionsetup[subfigure]{justification=centering}
  \begin{subfigure}{0.3\linewidth}
    \centering
    {\includegraphics[width=0.8\linewidth]{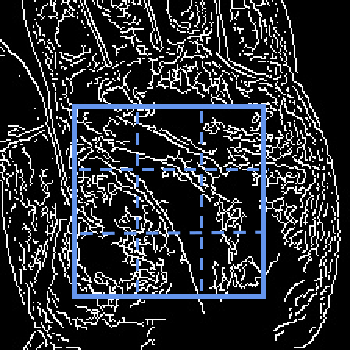}}
    \caption{$3 \times 3$ ROI blocks on the original background}
    \label{fig:systhesis_process-sub1}
  \end{subfigure}
  \begin{subfigure}{0.3\linewidth}
    \centering
    {\includegraphics[width=0.8\linewidth]{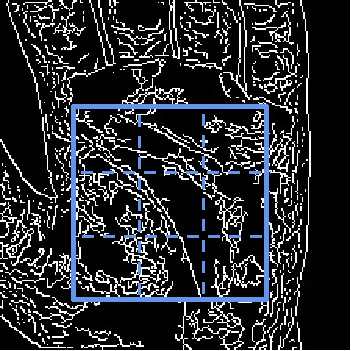}}
    \caption{ROI on a new background}
    \label{fig:systhesis_process-sub2}
  \end{subfigure}
  \begin{subfigure}{1.0\linewidth}
    \centering
    \includegraphics[width=1.0\linewidth]{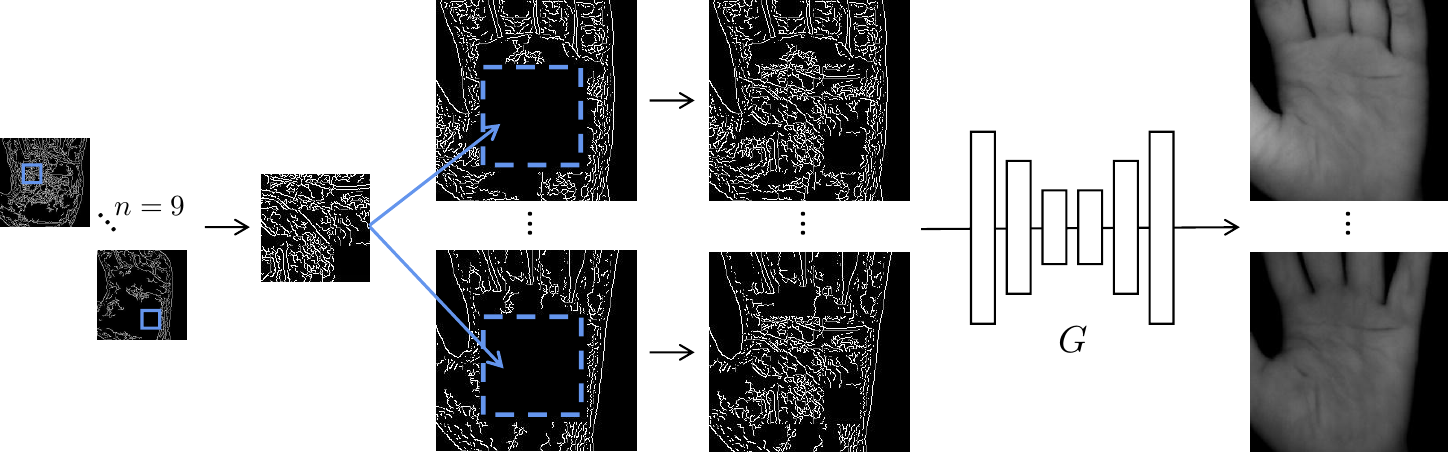}
    \caption{An example of the synthesis process}
    \label{fig:systhesis_process-sub3}
  \end{subfigure}
  \caption{The synthesis procedure of a new identity. (a) illustrates $3\times3$ blocks within the ROI. (b) shows the cropped ROI on a new background. (c) is an example of the synthesis process. By replacing the blocks with corresponding areas from nine identities, a unique cropped ROI can be determined. The generator generates an unseen identity from this unique input. By replacing the background outside the same ROI, we generate new samples of the same identity for different gestures.}
  \label{fig:systhesis_process}
\end{figure}

Let $\mathcal{I}=\{I_i, i=1,...,N\}$ be a set of Canny images corresponding to $N$ different identities.
Each $\{I_i\}$ consists of $3 \times 3$ blocks, numbered from top-left to bottom-right as $\{I_i^j, j=1,...,9\}$ as shown in \cref{fig:systhesis_process-sub1}. 
We can re-assemble a new ROI by selecting one block from a different $\{I_i\}$ for each location, for example, $\{I_1^1, I_2^2, ..., I_9^9\}$, and use the new combination to replace the corresponding region of a Canny image. 
The combination of blocks should adhere to their respective numbered areas in the original images to avoid violating physiological constraints.
The generator uses this new Canny image as the condition to generate an unseen palmprint image.
The synthesis process is illustrated in \cref{fig:systhesis_process-sub3}. 

As shown in \cref{fig:systhesis_process-sub2} and \cref{fig:systhesis_process-sub3}, when the ROI of a Canny image is determined, the background outside the ROI can be replaced to enrich the generation diversity.
We collect a wide range of diverse gestures from the same person, the gesture replacement within the same identity ensures intra-class diversity.

\subsection{Guidelines for Inter-class Distinction}
\label{inter-class-distinction}

To ensure significant inter-class distinction between any two combinations, we follow two guidelines when re-assembling new combinations:
\begin{enumerate}
    \item Only one block per identity is used in a combination.
    \item At least four different blocks are required between any two combinations.
\end{enumerate}

Given $N$ identities, it is important to determine the maximum number of new identities that can be synthesized under these guidelines.
This problem can be translated into a special case of the maximal clique problem in graph theory \cite{luce1949method}.
We can use a straightforward greedy algorithm to find a single maximal clique, which is actually a list of identity selections.

For simplicity, we use an approximated method to generate combinations, rather than enumerate every possible case.
Given a sample of nine identities $\{I_1, I_2, ..., I_9\}$, we generate block combinations following the circular shift, \eg, $\{I_1^1, I_2^2, ..., I_9^9\}$, $\{I_2^1, I_3^2, ..., I_1^9\}$, $\{I_3^1, I_4^2, ..., I_2^9\}$, \etc. With 30 identities, we can synthesize a maximum of 10,693 new identities, which is a considerable amount.

\section{Experiments}
\label{sec:experiments}

\subsection{Palmprint Datasets}
\label{sec:datasets}

We use eight public datasets in the experiments, the statistical information is summarized in \cref{tab:datasets}. The synthetic datasets NS-PVDB \cite{salazar2021automatic} and Synthetic-sPVDB \cite{salazar2021generating} are used in pre-training, and the remaining six datasets are employed for fine-tuning. We reprocess some fine-tuning datasets, including identity merging and ROI segmentation. The detailed descriptions of the reprocessed datasets are as follows:

\begin{table}[t]
  \caption{Statistics of palmprint datasets used in our experiments.
  }
  \label{tab:datasets}
  \centering
    \begin{tabular}{llll}
      \toprule
      Name                                        & \multicolumn{1}{c}{\#IDs}     & \multicolumn{1}{c}{\#Images}  & \multicolumn{1}{c}{Device} \\ 
      \midrule
      CASIA \cite{sun2005ordinal}                 & \multicolumn{1}{c}{620}       & \multicolumn{1}{c}{5,502}      & \multicolumn{1}{c}{Contactless}     \\
      CASIA-MS \cite{hao2008multispectral}        & \multicolumn{1}{c}{200}       & \multicolumn{1}{c}{2,400}      & \multicolumn{1}{c}{Contactless}     \\
      IITD \cite{kumar2008incorporating}          & \multicolumn{1}{c}{460}       & \multicolumn{1}{c}{2,601}      & \multicolumn{1}{c}{Contactless}     \\
      MPD \cite{zhang2020towards}                 & \multicolumn{1}{c}{400}       & \multicolumn{1}{c}{16,000}     & \multicolumn{1}{c}{Phone}           \\
      PolyU-V2 \cite{kanhangad2010contactless}    & \multicolumn{1}{c}{114}       & \multicolumn{1}{c}{570}       & \multicolumn{1}{c}{3D scanner}      \\
      TCD \cite{zhang2017towards}                 & \multicolumn{1}{c}{600}       & \multicolumn{1}{c}{12,000}     & \multicolumn{1}{c}{Contactless}  \\ \midrule
      NS-PVDB \cite{salazar2021automatic} & 16,000 & \multicolumn{1}{c}{112,000} & \multicolumn{1}{c}{N/A} \\
      Synthetic-sPVDB \cite{salazar2021generating} & 20,000 & \multicolumn{1}{c}{140,000} & \multicolumn{1}{c}{N/A} \\
      \bottomrule   
    \end{tabular}
\end{table}

\noindent\textbf{CASIA Multi-Spectral Palmprint Database (CASIA-MS)} \cite{hao2008multispectral} contains multi-spectral palm images, we only use images collected at 850nm wavelength to verify the performance of our method on palm vein.

\noindent\textbf{IIT Delhi Touchless Palmprint Database (IITD)} \cite{kumar2008incorporating} provides automatically segmented and normalized images. We reprocess the original images with hand key-points detection to obtain ROI images from the right perspective.

\noindent\textbf{Tongji Mobile Palmprint Dataset (MPD)} \cite{zhang2020towards} consists of images in two periods with various background and lighting conditions, which are captured by two brands of mobile phones. We merge these four categories under the same identity to verify the generalization performance, instead of removing the overlapped parts.

\noindent\textbf{Tongji Contactless Palmprint Database (TCD)} \cite{zhang2017towards} contains images collected from two different time periods with an average interval of approximately 61 days, named ``session1''
and ``session2'' in the dataset. To verify the stability of our method over time intervals, we merge images under the same identity from these two time periods.

In total, we have 2,394 identities which are divided into a training set and a test set under the \textit{open-set protocol}, wherein the identities present in the training set are not included in the test set. On the contrary, the \textit{closed-set protocol} means that identities in training set may appear in the test set, such as the samples at two different time periods in TCD dataset. We conduct closed-set experiments individually on each dataset and open-set experiments on the combined dataset under two different training-to-test ratios, 1:1 (1,197:1,197) and 1:3 (594:1,786), where 1:3 represents a more difficult real-world scenario.

We use three datasets in the pre-training phase for comparison purpose. In addition to the two synthetic datasets in \cref{tab:datasets}, we reproduce B$\rm\acute{e}$zierPalm \cite{zhao2022bezierpalm} with the open-source code.

\subsection{Implementation Details}

\noindent\textbf{Data Synthesis Settings}. Our implementation is based on the open-source Pix2Pix \cite{phillip2017pix2pixgit}. We use U-Net \cite{ronneberger2015u} as the generator and PatchGAN \cite{isola2017image} as the discriminator. The models are trained with Adam \cite{kingma2014adam} optimizer with the initial learning rate as 2e-4 and $\beta_1 = 0.5$. For the loss function configuration, we employ the mean squared error of LSGAN \cite{mao2017least} as the discriminator loss, the balancing factor $\lambda$ in \cref{eq:loss_final} is set to 100. The models are fed with Canny edge images of normalized palms, with a fixed size of 256$\times$256. We train the models on our internal dataset, which contains 100,000 samples of 1,000 identities captured by contactless camera. All the images are normalized by hand key-points illustrated in \cref{fig:alignment_canny} and processed by Canny edge detector. The high-threshold and low-threshold of Canny algorithm are set to 30 and 5, respectively. For synthetic dataset, we generate 10,000 identities, the handedness is randomly changed by flipping. Each identity contains 500 samples with various gestures and a total of 5 million images are used for pre-training.

\noindent\textbf{Recognition Training Settings}. 
We employ ArcFace \cite{deng2019arcface}, a face recognition framework, for the palmprint recognition because of the similarity between these two tasks. 
We pre-train the model on synthesized data and then fine-tune on public palmprint datasets. The input size of the image in both phases is 224$\times$224. We employ augmentations including brightness, contrast, rotation and motion blur, and additionally add border cutout in pre-training. In the ArcFace framework, the margin $m$ is 0.5 and the scale factor $s$ is 48, the final embedding feature $x\in \mathbb{R}^{512}$. The ArcFace baseline is directly trained on public datasets without pre-training. We use the cosine annealing learning rate scheduler with half cycle set to 1 epoch, the learning rate for pre-training and fine-tuning is 0.1 and 0.01, respectively. In fine-tuning phase, we apply warmup start for learning rate if the model has pre-training weights. We use SGD algorithm as the optimizer with weight decay and momentum set to 5e-4 and 0.9, respectively. The batch size is 128 and the dropout for model is set to 0.2.

\subsection{Evaluation Results}

We adopt average top-1 accuracy and equal error rate (EER) as evaluation metrics. To simulate the real user experience, we randomly choose one image in each identity as the registry, and the remaining images serve as queries. A query will calculate the distance from all registries, and it is considered a successful match only when the query is the closest to the registry of the same identity. The calculation of top-1 accuracy is based on the registry-query rule, and EER is the point where false acceptance rate (FAR) and false rejection rate (FRR) intersect in ROC curve.

\noindent\textbf{Closed-set Palmprint Recognition}. Closed-set experiments are conducted on four datasets with a relatively large number of identities: CASIA, IITD, TCD and MPD. We employ 5-fold cross-validation to evaluate MobileFaceNet performance,
the test results are presented in \cref{tab:closed-set}, including ArcFace baseline without pre-training and four pre-training methods. While the top-1 accuracy results under the closed-set protocol are approaching saturation, our method significantly reduce the magnitude of EER, which indicates that our model exhibits better performance under stricter FAR requirements.

\begin{table*}
  \caption{Top-1 accuracy(\%) and EER(\%) results of closed-set recognition.}
  \label{tab:closed-set}
  \centering
  \begin{tabular}{lllll}
  \toprule
  Method           & \multicolumn{1}{c}{CASIA} & \multicolumn{1}{c}{IITD} & \multicolumn{1}{c}{TCD} & \multicolumn{1}{c}{MPD} \\ 
  \midrule
  No pre-training      & 99.31 / 0.589 & 98.41 / 1.320 & 99.54 / 0.412 & 96.96 / 1.663    \\
  Synthetic-sPVDB  & 99.16 / 0.654 & 98.93 / 0.929 & 99.34 / 0.484 & 97.96 / 1.403 \\
  NS-PVDB  & 99.60 / 0.362 & 99.39 / 0.561 & 99.82 / 0.294 & 98.54 / 1.168 \\ 
  B$\rm\acute{e}$zierPalm  & 99.77 / 0.256   &  99.44 / 0.522  & 100.0 / 0.128  & 99.08 / 0.769    \\
  Ours & \textBF{99.92} / \textBF{0.174}   & \textBF{100.0} / \textBF{0.056}  & \textBF{100.0} / \textBF{0.059}  & \textBF{99.96} / \textBF{0.135} \\
  \bottomrule
  \end{tabular}
\end{table*}

\noindent\textbf{Open-set Palmprint Recognition}. The open-set test using the combined datasets provides a comprehensive evaluation of the model's identification and generalization performance. We set up two training-to-test ratios and the quantitative results are shown in \cref{tab:open-set}.

Under open-set protocol, the ArcFace baseline trained from scratch exhibits lower identification performance, which further validates the importance of pre-training.
Our proposed method significantly improves the baseline under both 1:1 and 1:3 ratios, with the improvement in TAR increasing as the FAR metrics become more stringent. Canny2Palm outperforms other methods under all settings with more photorealistic data and more controllable synthesis pipeline.

\begin{table*}
    \caption{Open-set test results. ``MFN'' refers to MobileFaceNet and ``R50'' refers to ResNet50.}
    \label{tab:open-set}
    \centering
    \begin{tabular}{ll|llll|llll}
    \toprule
    \multirow{3}{*}{Method} & \multirow{3}{*}{Backbone} & \multicolumn{4}{c|}{train:test = 1:1} & \multicolumn{4}{c}{train:test = 1:3} \\
                                &                           & \multicolumn{4}{c|}{TAR@FAR=}         & \multicolumn{4}{c}{TAR@FAR=}         \\
                                &                           & \multicolumn{1}{c}{1e-3}    & \multicolumn{1}{c}{1e-4}   & \multicolumn{1}{c}{1e-5}   & \multicolumn{1}{c|}{1e-6}   &
                                \multicolumn{1}{c}{1e-3}    & \multicolumn{1}{c}{1e-4}    & \multicolumn{1}{c}{1e-5}   & \multicolumn{1}{c}{1e-6}   \\ 
    \midrule
    No pre-training & \multicolumn{1}{c|}{MFN}                    & 88.28   & 79.64  & 70.77  & 61.17  & 83.72   & 72.13  & 60.05  & 46.70  \\
    Synthetic-sPVDB  & \multicolumn{1}{c|}{MFN}                & 90.05   & 82.53 & 75.01 & 67.54 & 83.32   &  74.26  &  65.69  &  59.07  \\
    NS-PVDB  & \multicolumn{1}{c|}{MFN}                & 90.77        & 84.30       & 77.01       &  69.50      &  86.16  &  78.37 &  70.59 & 64.34 \\
    B$\rm\acute{e}$zierPalm  & \multicolumn{1}{c|}{MFN} & 92.03   & 85.78  & 79.15  & 73.11  & 90.31   & 82.82  & 73.97  & 65.11  \\
    Ours & \multicolumn{1}{c|}{MFN}               & \textBF{94.51}     & \textBF{90.55}    & \textBF{85.62}   & \textBF{80.31}  &
                                                                                        \textBF{91.50}  & \textBF{85.02}  & \textBF{77.63} & \textBF{69.88}  \\ \hline
    No pre-training & \multicolumn{1}{c|}{R50}  & 90.66   &  81.31   & 69.35  & 57.60  & 75.93   & 24.58    & \multicolumn{1}{r}{8.60}  & \multicolumn{1}{r}{3.20}  \\        
    Synthetic-sPVDB  & \multicolumn{1}{c|}{R50}                & 93.27       & 86.95    & 78.37    &  70.68     & 88.04   &  78.54  & 68.68   &  60.79  \\                                                            
    NS-PVDB  & \multicolumn{1}{c|}{R50}   & 95.13        & 90.35       &  84.32      & 77.24       &  93.28       & 86.29       &  77.87      &   67.90     \\
    B$\rm\acute{e}$zierPalm  & \multicolumn{1}{c|}{R50}  & 95.49        & 90.86       &  84.92      & 79.50       & 93.50  &  86.21  & 77.26  & 66.21 \\
    Ours &\multicolumn{1}{c|}{R50}  &  \textBF{95.93}       &  \textBF{91.49}      &  \textBF{86.23}      &  \textBF{80.99}      & \textBF{93.72}  & \textBF{86.45}  &  \textBF{77.92} & \textBF{70.15} \\
    \bottomrule
    \end{tabular}
\end{table*}

\noindent\textbf{Cross-dataset Validation}. One concern in deep learning-based palmprint recognition is the issue of overfitting. To measure the generalization capability of the proposed method, we conduct cross-dataset validation under four different settings. The performance is evaluated by top-1, EER and TAR@FAR indicators using MobileFaceNet backbone, which is presented in \cref{tab:cross-datasets}. 

Compared to ArcFace baseline, our method consistently improves the performance under all settings, particularly when considering the indicator of FAR=1e-5. For example, when training on the grayscale dataset TCD and testing on the RGB dataset IITD, the performance of TAR@(FAR=1e-5) is improved by approximately 30\%, which demonstrates the strong generalization capability of the pre-trained model in cross-dataset scenarios.

\begin{table}[t]
  \caption{Cross-dataset validation results. ``T'', ``I'', ``C'' and ``M'' represent TCD, IITD, CASIA and MPD dataset, respectively. T$\rightarrow$I indicates that the model is trained on TCD dataset and evaluated on IITD dataset.}
  \label{tab:cross-datasets}
  \centering
  \begin{tabular}{ll|ll|lll}
  \toprule
  \multirow{2}{*}{Datasets} & \multirow{2}{*}{Method} & \multirow{2}{*}{Top-1} & \multirow{2}{*}{EER} & \multicolumn{3}{c}{TAR@FAR=} \\
                        &                         &                        &                      & 1e-3     & 1e-4    & 1e-5    \\ 
  \midrule
  \multirow{2}{*}{T$\rightarrow$I}   & No pre-training        &  91.87                 &  3.503               & 88.74    & 82.91   & 63.90   \\
                                  & Ours           &  \textBF{99.44}        &  \textBF{0.467}      & \textBF{98.46}    & \textBF{96.31}   & \textBF{93.65}   \\ \hline
  \multirow{2}{*}{T$\rightarrow$C}   & No pre-training        &  97.64                 &  1.380               & 96.07    & 93.37   & 89.12   \\
                                  & Ours           &  \textBF{99.06}        &  \textBF{0.815}      & \textBF{98.62}    & \textBF{97.03}   & \textBF{95.82}   \\ \hline
  \multirow{2}{*}{M$\rightarrow$I}   & No pre-training        &  98.88                 &  0.812               & 98.32    & 95.89   & 92.90   \\
                                  & Ours           &  \textBF{99.77}        &  \textBF{0.424}      & \textBF{99.49}    & \textBF{98.09}   & \textBF{96.82}   \\ \hline
  \multirow{2}{*}{M$\rightarrow$C}   & No pre-training        &  99.18                 &  0.626               & 98.81    & 97.39   & 92.91   \\
                                  & Ours           &  \textBF{99.67}        &  \textBF{0.242}      & \textBF{99.58}    & \textBF{98.87}   & \textBF{97.99}  \\
  \bottomrule
  \end{tabular}
\end{table}

\subsection{Evaluation at Million Scale}

To further validate the generalization capability of our proposed method, we collect a million-level internal dataset to evaluate the performance. The training dataset contains 6,000 identities and 27.3 million samples, while the test dataset contains 700 identities and 2.2 million samples. The number of inter-class comparison pairs reaches 1.5 billion. We compare our pre-trained model with the ArcFace baseline, which is trained from scratch. The evaluation is under open-set protocol and the FAR metric is set from 1e-5 to 1e-8. As shown in \cref{tab:million-scale}, the proposed method outperforms the baseline under all metrics. The improvement increases when switching the model from R50 to MFN and testing under a more strict FAR.

\begin{figure}
  \begin{minipage}[t]{.45\linewidth}
  \captionof{table}{Evaluation results on an internal million scale dataset.}
  \label{tab:million-scale}
  \centering
  \resizebox{\linewidth}{!}{
  \begin{tabular}{ll|llll}
  \toprule
  \multirow{2}{*}{Method} & \multirow{2}{*}{Backbone}  & \multicolumn{4}{c}{TAR@FAR=} \\
                    &          &     1e-5       &    1e-6   & 1e-7     & 1e-8     \\ 
  \midrule
  No pre-training  & \multicolumn{1}{c|}{\multirow{2}{*}{MFN}}     &  95.35               & 94.18    & 93.13   & 91.77   \\
  Ours     &             &  \textBF{95.90}               & \textBF{94.99}    & \textBF{94.03}   & \textBF{92.73}   \\ \hline
  No pre-training  &  \multicolumn{1}{c|}{\multirow{2}{*}{R50}}     &   97.83      &  96.76   &  95.74  & 94.21   \\
  Ours        &     &  \textBF{98.42}     & \textBF{97.25}    & \textBF{96.37}   & \textBF{94.91}   \\
  \bottomrule
  \end{tabular}
  }
  \end{minipage}
  \hfill
  \begin{minipage}[t]{.5\linewidth}
  \captionof{table}{Ablation study on augmentation design. ``C'' and ``B'' represent border cutout and random backgrounds, respectively.}
  \label{tab:design-choices}
  \centering
  \begin{tabular}{ll|llll}
  \toprule
  \multirow{2}{*}{C} & \multicolumn{1}{c|}{\multirow{2}{*}{B}}  & \multicolumn{4}{c}{TAR@FAR=} \\
                     &          &     1e-3       &    1e-4   & 1e-5     & 1e-6     \\ 
  \midrule
  \multicolumn{2}{c|}{baseline}       &  81.91               & 72.56    & 64.37   & 56.36   \\
  \cmark            &    \multicolumn{1}{c|}{\xmark}         &  82.97             & 73.75    & 65.51   & 58.54   \\
  \xmark            &   \multicolumn{1}{c|}{\cmark}     &  93.19               & 88.04    & 82.04   & 73.57   \\
  \cmark        &   \multicolumn{1}{c|}{\cmark}   &  \textBF{94.51}     & \textBF{90.55}    & \textBF{85.62}   & \textBF{80.31}   \\
  \bottomrule
  \end{tabular}
  \end{minipage}
\end{figure}

\subsection{Ablation Study}

In this section, we explore different settings in Canny2Palm. All experiments are conducted with MobileFaceNet and evaluated under 1:1 open-set protocol.

\noindent
\textbf{Augmentation design}. We first ablate the augmentation design in pre-training, including the application of random backgrounds during the synthesis and the border cutout in pre-training. \cref{tab:design-choices} presents the test results with and without these designs. Under the metric of FAR=1e-6, the addition of random backgrounds significantly increases the intra-class diversity and improves the baseline by approximately 17.2\%. While the improvement of the border cutout is limited without background changes. However, the application of the border cutout based on random backgrounds data further improves the performance by approximately 5.7\%. We infer that under this design, the border cutout compels the model to prioritize the extraction of the ROI texture, thus effectively eliminating the overfitting from backgrounds.

\begin{figure}
  \centering
  \begin{subfigure}{0.47\linewidth}
    \includegraphics[width=1.0\linewidth]{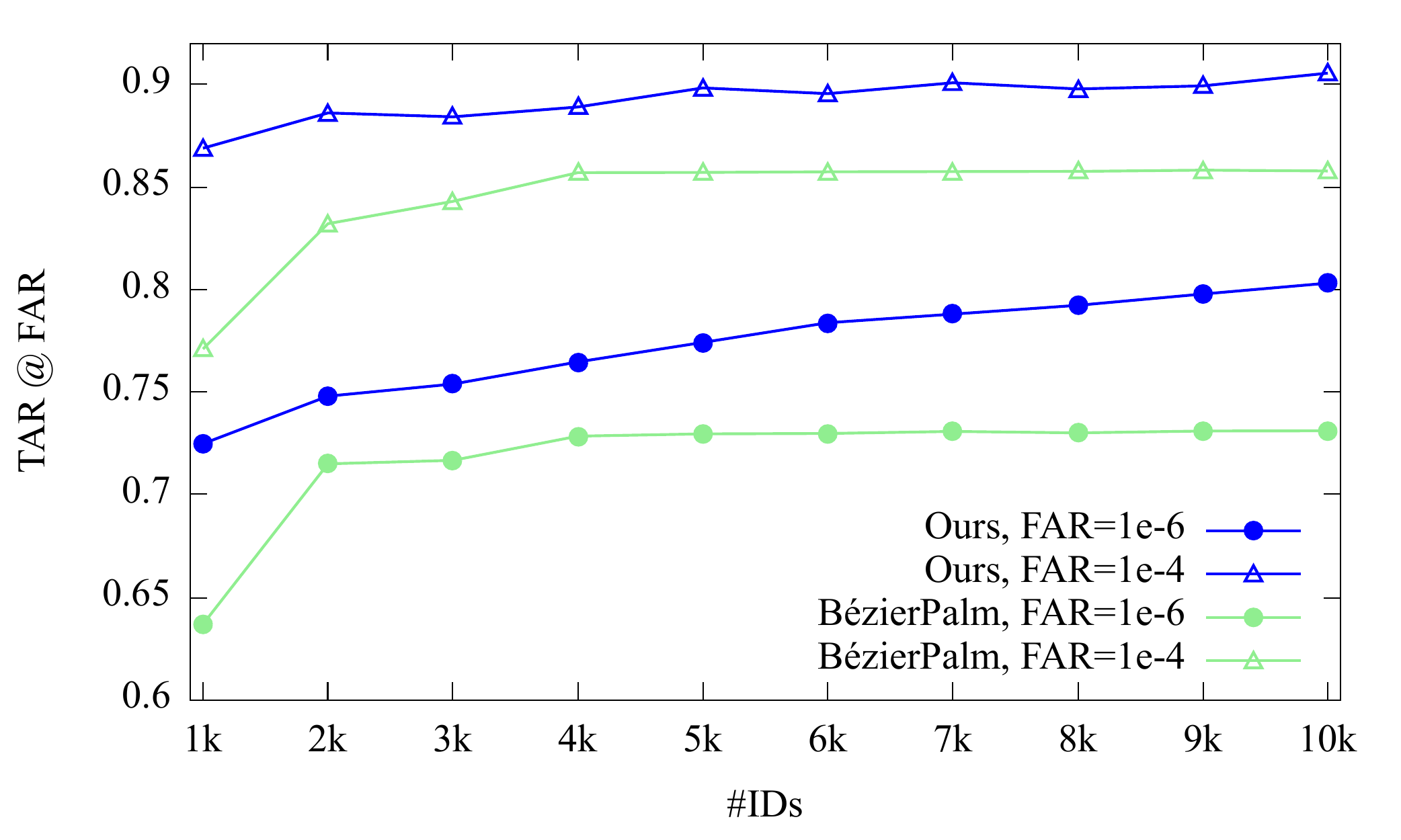}
    \caption{Number of identities}
    \label{fig:ablation-identity}
  \end{subfigure}
  \begin{subfigure}{0.47\linewidth}
    \includegraphics[width=1.0\linewidth]{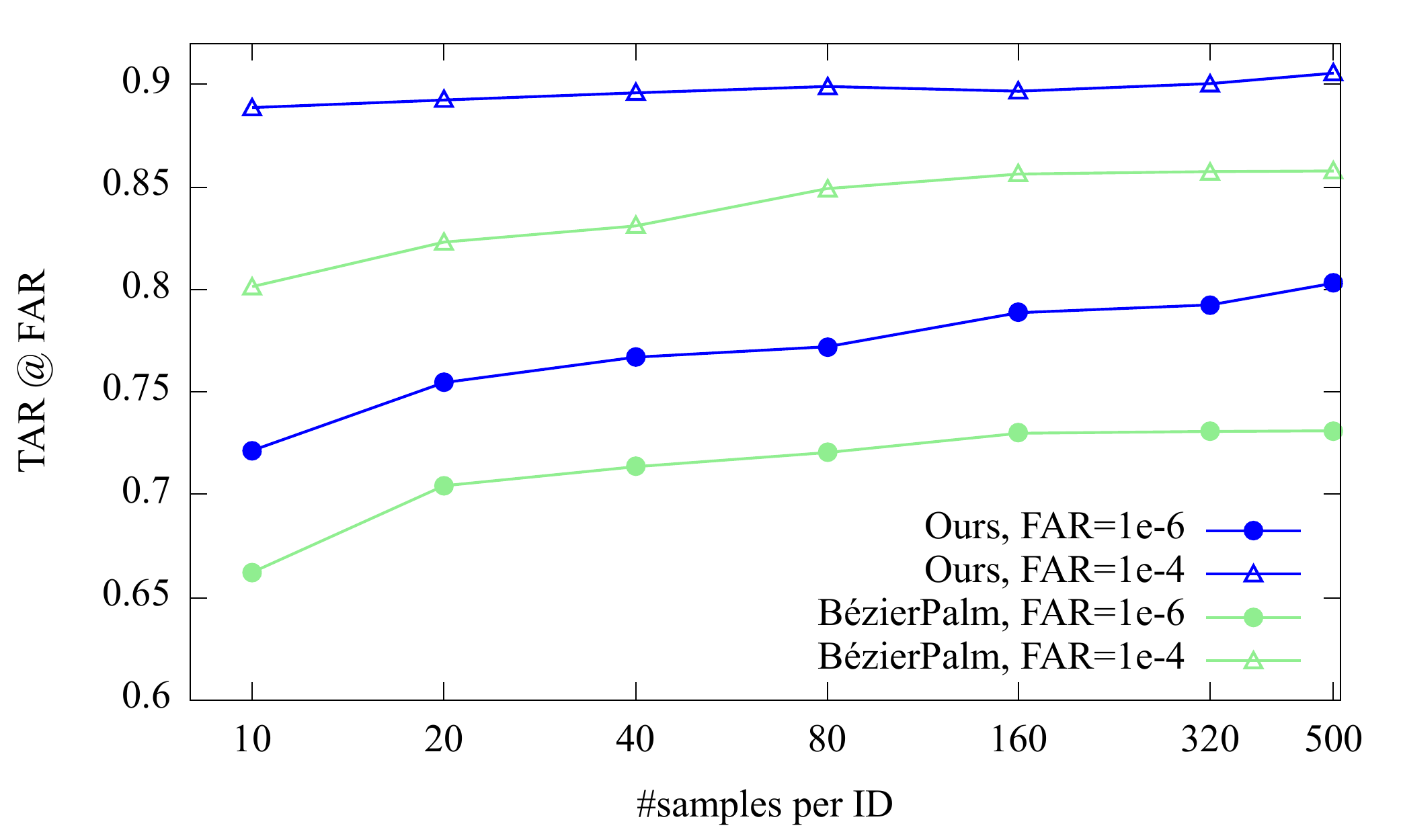}
    \caption{Number of samples}
    \label{fig:ablation-samples}
  \end{subfigure}
  \caption{Ablation test results with different synthesized identities and samples.}
  \label{fig:ablation-number}
\end{figure}

\noindent\textbf{Number of synthesized identities}. 
We synthesize 10,000 identities to evaluate the recognition performance with respect to the size of the pre-training data.
The number of samples per identity is fixed at 500.
\cref{fig:ablation-identity} shows the TAR accuracy curves for FAR=1e-4 and FAR=1e-6.
TAR tends to saturate more slowly under more strict FAR.
In comparison to B$\rm\acute{e}$zierPalm, models pre-trained with Canny2Palm not only achieve higher accuracy at the same number of identities, but also keep improving given more identities.
The performance of models pre-trained with B$\rm\acute{e}$zierPalm saturates around 4,000 identities.

\noindent\textbf{Number of samples per identity}.
In this experiment, we fix the number of identities to 10,000 and adjust the sample size of each identity. The test results in \cref{fig:ablation-samples} reveal that the sample size has a positive impact on the accuracy. This result further confirms the importance of intra-class diversity, which indicates the advantage of using samples with different gestures as backgrounds in our proposed method.
Similar to the saturation phenomenon witnessed in the previous ablation study, the performance curve of Canny2Palm flattens out slower than the counterpart.

\section{Conclusion}
\label{sec:conclusion}

In this paper, we propose Canny2Palm, a palmprint synthesis method to generate data for palmprint recognition pretraining.
Canny2Palm generates realistic palmprints with controllable intra-class diversity and inter-class distinction.
Canny2Palm can synthesize new identities in large scale without extra uniqueness classifier.
Pre-training palmprint recognition models with Canny2Palm synthetic data enhances the model generalization ability and mitigates overfitting.
Comprehensive experiments consistently demonstrate that our method outperforms state-of-the-art methods, thereby offering an effective solution to address the data scarcity problem in palmprint recognition.

\bibliographystyle{splncs04}
\bibliography{main}

\begin{thebibliography}{10}
\providecommand{\url}[1]{\texttt{#1}}
\providecommand{\urlprefix}{URL }
\providecommand{\doi}[1]{https://doi.org/#1}

\bibitem{canny1986computational}
Canny, J.: A computational approach to edge detection. IEEE Transactions on pattern analysis and machine intelligence (6),  679--698 (1986)

\bibitem{deng2019arcface}
Deng, J., Guo, J., Xue, N., Zafeiriou, S.: Arcface: Additive angular margin loss for deep face recognition. In: Proceedings of the IEEE/CVF conference on computer vision and pattern recognition. pp. 4690--4699 (2019)

\bibitem{goodfellow2014generative}
Goodfellow, I., Pouget-Abadie, J., Mirza, M., Xu, B., Warde-Farley, D., Ozair, S., Courville, A., Bengio, Y.: Generative adversarial nets. Advances in neural information processing systems  \textbf{27} (2014)

\bibitem{hao2008multispectral}
Hao, Y., Sun, Z., Tan, T., Ren, C.: Multispectral palm image fusion for accurate contact-free palmprint recognition. In: 2008 15th IEEE International Conference on Image Processing. pp. 281--284. IEEE (2008)

\bibitem{isola2017image}
Isola, P., Zhu, J.Y., Zhou, T., Efros, A.A.: Image-to-image translation with conditional adversarial networks. In: Proceedings of the IEEE conference on computer vision and pattern recognition. pp. 1125--1134 (2017)

\bibitem{kanhangad2010contactless}
Kanhangad, V., Kumar, A., Zhang, D.: Contactless and pose invariant biometric identification using hand surface. IEEE transactions on image processing  \textbf{20}(5),  1415--1424 (2010)

\bibitem{kingma2014adam}
Kingma, D.P., Ba, J.: Adam: A method for stochastic optimization. arXiv preprint arXiv:1412.6980  (2014)

\bibitem{kingma2013auto}
Kingma, D.P., Welling, M.: Auto-encoding variational bayes. arXiv preprint arXiv:1312.6114  (2013)

\bibitem{kumar2008incorporating}
Kumar, A.: Incorporating cohort information for reliable palmprint authentication. In: 2008 Sixth Indian conference on computer vision, graphics \& image processing. pp. 583--590. IEEE (2008)

\bibitem{kumar2018toward}
Kumar, A.: Toward more accurate matching of contactless palmprint images under less constrained environments. IEEE Transactions on Information Forensics and Security  \textbf{14}(1),  34--47 (2018)

\bibitem{luce1949method}
Luce, R.D., Perry, A.D.: A method of matrix analysis of group structure. Psychometrika  \textbf{14}(2),  95--116 (1949)

\bibitem{mao2017least}
Mao, X., Li, Q., Xie, H., Lau, R.Y., Wang, Z., Paul~Smolley, S.: Least squares generative adversarial networks. In: Proceedings of the IEEE international conference on computer vision. pp. 2794--2802 (2017)

\bibitem{mirza2014conditional}
Mirza, M., Osindero, S.: Conditional generative adversarial nets. arXiv preprint arXiv:1411.1784  (2014)

\bibitem{phillip2017pix2pixgit}
Phillip, I., Jun-Yan, Z., Tinghui, Z., Alexei~A., E.: Cyclegan and pix2pix in pytorch. \url{https://github.com/junyanz/pytorch-CycleGAN-and-pix2pix} (2017)

\bibitem{ronneberger2015u}
Ronneberger, O., Fischer, P., Brox, T.: U-net: Convolutional networks for biomedical image segmentation. In: Medical Image Computing and Computer-Assisted Intervention--MICCAI 2015: 18th International Conference, Munich, Germany, October 5-9, 2015, Proceedings, Part III 18. pp. 234--241. Springer (2015)

\bibitem{salazar2021automatic}
Salazar, E., Hern{\'a}ndez-Garc{\'\i}a, R., Barrientos, R., Vilches, K., Mora, M., V{\'a}squez, A.: Automatic generation of synthetic palm vein images: a nature-based approach. In: 11th International Conference of Pattern Recognition Systems (ICPRS 2021). vol.~2021, pp. 38--43. IET (2021)

\bibitem{salazar2021generating}
Salazar, E., Hern{\'a}ndez-Garc{\'\i}a, R., Barrientos, R., Vilches, K., Mora, M., V{\'a}squez, A.: Generating style-based palm vein synthetic images for the creation of large-scale datasets. In: 11th International Conference of Pattern Recognition Systems (ICPRS 2021). vol.~2021, pp. 182--187. IET (2021)

\bibitem{salazar2022towards}
Salazar-Jurado, E.H., Hern{\'a}ndez-Garc{\'\i}a, R., Vilches-Ponce, K., Barrientos, R.J., Mora, M., Jaswal, G.: Towards the generation of synthetic images of palm vein patterns: a review. arXiv preprint arXiv:2205.10179  (2022)

\bibitem{shen2023rpg}
Shen, L., Jin, J., Zhang, R., Li, H., Zhao, K., Zhang, Y., Zhang, J., Ding, S., Zhao, Y., Jia, W.: Rpg-palm: Realistic pseudo-data generation for palmprint recognition. In: Proceedings of the IEEE/CVF International Conference on Computer Vision. pp. 19605--19616 (2023)

\bibitem{sun2005ordinal}
Sun, Z., Tan, T., Wang, Y., Li, S.Z.: Ordinal palmprint represention for personal identification [represention read representation]. In: 2005 IEEE computer society conference on computer vision and pattern recognition (CVPR'05). vol.~1, pp. 279--284. IEEE (2005)

\bibitem{zhang2017towards}
Zhang, L., Li, L., Yang, A., Shen, Y., Yang, M.: Towards contactless palmprint recognition: A novel device, a new benchmark, and a collaborative representation based identification approach. Pattern Recognition  \textbf{69},  199--212 (2017)

\bibitem{zhang2019pay}
Zhang, Y., Zhang, L., Liu, X., Zhao, S., Shen, Y., Yang, Y.: Pay by showing your palm: A study of palmprint verification on mobile platforms. In: 2019 IEEE International Conference on Multimedia and Expo (ICME). pp. 862--867. IEEE (2019)

\bibitem{zhang2020towards}
Zhang, Y., Zhang, L., Zhang, R., Li, S., Li, J., Huang, F.: Towards palmprint verification on smartphones. arXiv preprint arXiv:2003.13266  (2020)

\bibitem{zhao2022bezierpalm}
Zhao, K., Shen, L., Zhang, Y., Zhou, C., Wang, T., Zhang, R., Ding, S., Jia, W., Shen, W.: B{\'e}zierpalm: A free lunch for palmprint recognition. In: European Conference on Computer Vision. pp. 19--36. Springer (2022)

\bibitem{zhong2019decade}
Zhong, D., Du, X., Zhong, K.: Decade progress of palmprint recognition: A brief survey. Neurocomputing  \textbf{328},  16--28 (2019)

\bibitem{zhu2021webface260m}
Zhu, Z., Huang, G., Deng, J., Ye, Y., Huang, J., Chen, X., Zhu, J., Yang, T., Lu, J., Du, D., et~al.: Webface260m: A benchmark unveiling the power of million-scale deep face recognition. In: Proceedings of the IEEE/CVF Conference on Computer Vision and Pattern Recognition. pp. 10492--10502 (2021)

\end{thebibliography}

\appendix

\section{Appendix}

\subsection{Algorithm for finding a single maximal clique}

\begin{figure}
  \centering  
  \includegraphics[width=0.6\linewidth]{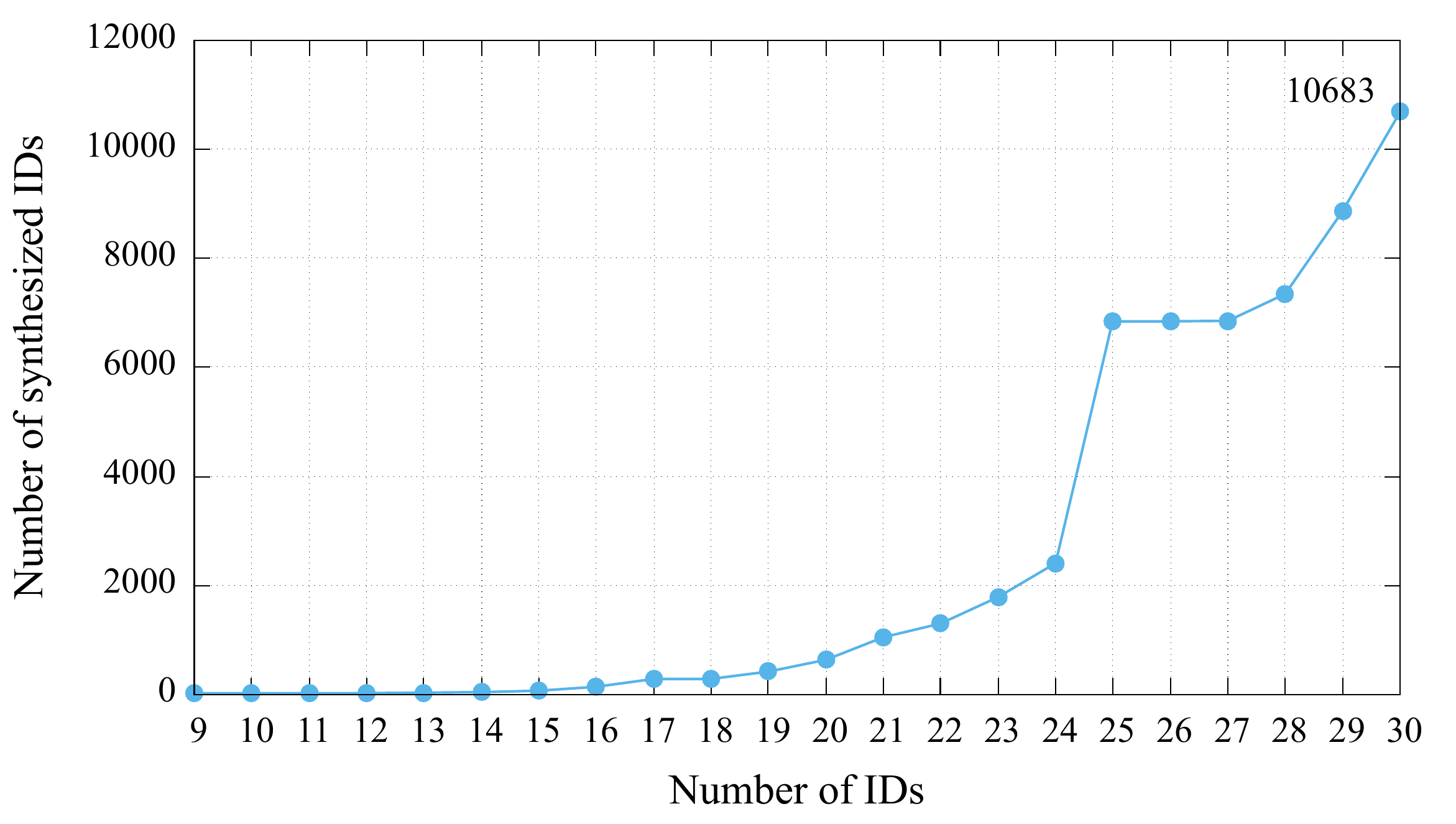}
  \caption{Maximum number of virtual identities that can be synthesized based on the given number of identities.}
  \label{fig:combination_numbers}
\end{figure}

In section 3.3 we employ a straightforward greedy algorithm to find a single maximal clique.
The algorithm is described in \cref{alg:greedy}. A complete synthesis process requires a combination of nine identities to obtain nine new identities under the guidelines. Here we plot the number of identities that can be synthesized ranging from 9 to 30 identities in \cref{fig:combination_numbers}.

\begin{figure}
  \centering  
  \captionsetup[subfigure]{justification=centering}
  \begin{subfigure}{0.2\linewidth}
    \centering
    \includegraphics[width=0.8\linewidth]{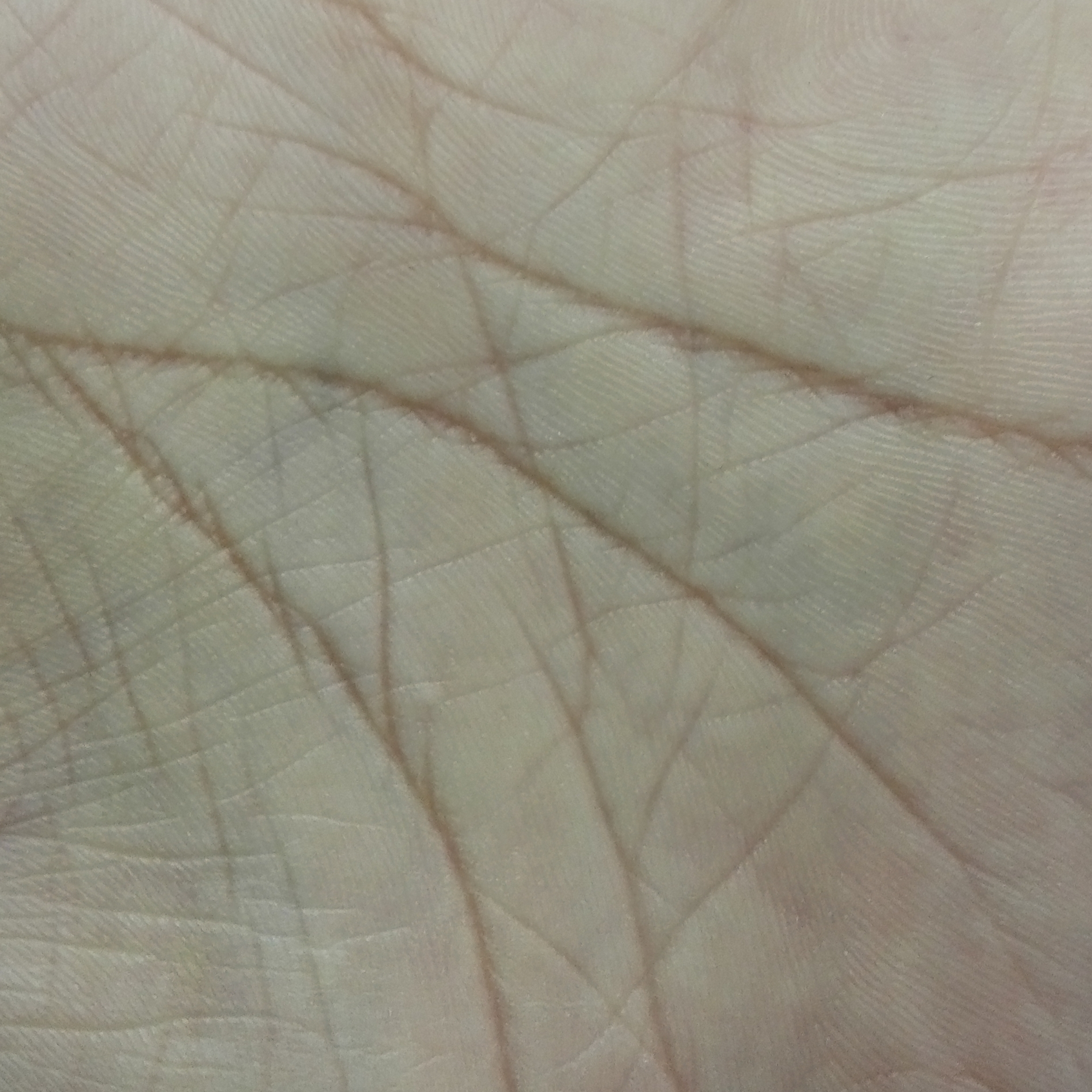}
    \caption{ROI example \\ from MPD}
    \label{fig:data_example-sub1}
  \end{subfigure}
  \begin{subfigure}{0.2\linewidth}
    \centering
    \includegraphics[width=0.8\linewidth]{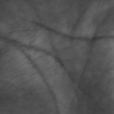}
    \caption{ROI example \\ from TCD}
    \label{fig:data_example-sub2}
  \end{subfigure}
  \begin{subfigure}{0.2\linewidth}
    \centering
    {\includegraphics[width=0.8\linewidth]{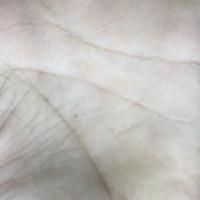}}
    \caption{Pre-processed \\ ROI from IITD}
    \label{fig:data_example-sub3}
  \end{subfigure}
  \begin{subfigure}{0.2\linewidth}
    \centering
    {\includegraphics[width=0.8\linewidth]{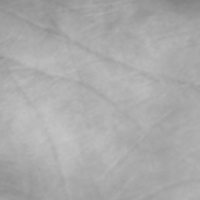}}
    \caption{Pre-processed \\ ROI from CASIA}
    \label{fig:data_example-sub4}
  \end{subfigure}
  \caption{Example ROI images of different datasets. (a) and (b) are segmented based on the finger-gap-point detection; (c) and (d) are pre-processed ROI images based on our hand key-points normalization.}
  \label{fig:data_example}
\end{figure}

\begin{algorithm}
  \caption{Python Pseudo-code for greedy algorithm to find a list of identity combinations.}
  \label{alg:greedy}
  \KwIn{Number of identities $N$, length of a combination $M$, the maximum number of duplicate identities for two combinations $K$, list of total combinations $T$}
  \KwOut{A list of identity combinations $L$ representing new virtual identities}
  initialize $L$ to an empty list, let $M=9, K=5$\;
  $T$ = itertools.combinations(range($N$), $M$), len($T$) = $C_N^M$\;
  \SetKwFunction{FMain}{AddVertex}
  \SetKwProg{Fn}{Function}{:}{}
  \Fn{\FMain{$vertex, clique$}}{
    \For{$j\leftarrow$ \rm len($clique$) \KwTo $1$}{
      $n_{duplicate}=2\times M-$ len(set($vertex+clique_j$))\;
      \eIf{$n_{duplicate}\leq K$}{Flag = True\;}{Flag = False\; \textbf{Break} } 
    }
    \textbf{return} Flag
  }
  \textbf{End Function}\;
  \For{$i\leftarrow 1$ \KwTo $C_N^M$}{
    \eIf{$i=1$}{
      $L$.append($T_i$)\;
    }{
      Flag = \FMain{$T_i$, $L$}\;
      \If{\rm Flag = True}{$L$.append($T_i$)\;}
    }
  }
\end{algorithm}

\subsection{Examples of reprocessed open source datasets}

In section 4.1, we reprocess IITD and CASIA datasets as they did not provide ROI segmentation. In most public palmprint datasets, the segmentation of ROI image is based on the finger-gap-point detection\cite{zhang2019pay}. While our normalization method includes redundant finger region information. To maintain consistency, we can simply segment the central region of our normalized image. The example ROI images are shown in \cref{fig:data_example}.

\subsection{Border cutout}

\begin{figure}
  \centering  
  \captionsetup[subfigure]{justification=centering}
  \begin{subfigure}{0.24\linewidth}
    \centering
    \includegraphics[width=0.82\linewidth]{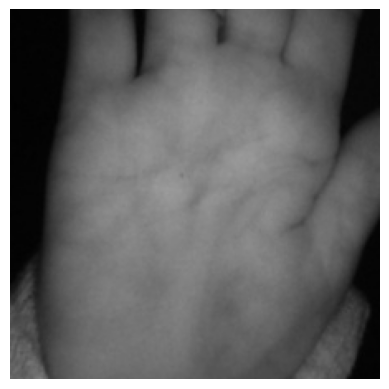}
    \caption{Original image}
    \label{fig:cutout-sub1}
  \end{subfigure}
  \begin{subfigure}{0.24\linewidth}
    \centering
    \includegraphics[width=0.82\linewidth]{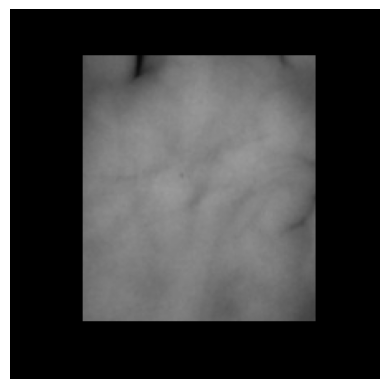}
    \caption{Apply cutout}
    \label{fig:cutout-sub2}
  \end{subfigure}
  \begin{subfigure}{0.24\linewidth}
    \centering
    \includegraphics[width=0.82\linewidth]{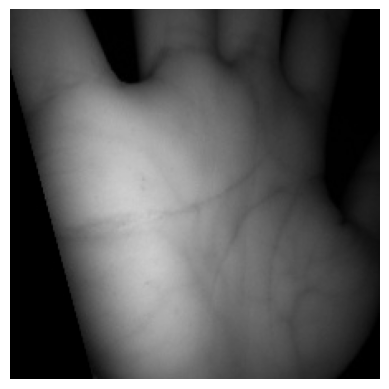}
    \caption{Original image}
    \label{fig:cutout-sub3}
  \end{subfigure}
  \begin{subfigure}{0.24\linewidth}
    \centering
    \includegraphics[width=0.82\linewidth]{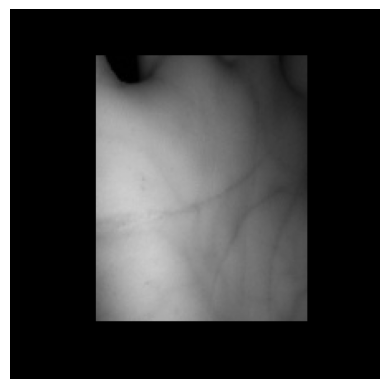}
    \caption{Apply cutout}
    \label{fig:cutout-sub4}
  \end{subfigure}
  \caption{Example images of border cutout augmentation.}
  \label{fig:cutout}
\end{figure}

We introduce an augmentation called border cutout to further improve the model performance in pre-training. As illustrated in \cref{fig:cutout}, we apply cutout to four borders of the image. For one border, the largest cutout area is 1/4 of the image size. Through this augmentation, we eliminate redundant background information under normal gestures. At the same time, retaining the original image in pre-training also increases the generalization performance for difficult gestures, as shown in \cref{fig:cutout-sub3}, the palm with a large inclination is difficult to locate the ROI area.

\end{document}